\documentclass[runningheads]{llncs}
\usepackage{graphicx}

\usepackage{tikz}
\usepackage{comment} 
\usepackage{amsmath,amssymb} 
\usepackage{color}
\usepackage[pagebackref=true,breaklinks=true,letterpaper=true,colorlinks,bookmarks=false]{hyperref}

\usepackage[width=122mm,left=12mm,paperwidth=146mm,height=193mm,top=12mm,paperheight=217mm]{geometry}

\begin{document}
\pagestyle{headings}
\mainmatter
\def\ECCVSubNumber{1302}  

\title{Hierarchical Style-based Networks \\ for Motion Synthesis} 

\titlerunning{Style-based Motion Synthesis}
%
\author{
Jingwei Xu\inst{1}\thanks{denotes equal contribution.},
Huazhe Xu\inst{2}$^\star$,
Bingbing Ni\inst{1}\thanks{corresponding author.},
Xiaokang Yang\inst{1}, \\
Xiaolong Wang\inst{3},
Trevor Darrell\inst{2}
}
\authorrunning{Xu et al.}
%
\institute{Shanghai Jiao Tong University \and
University of California, Berkeley \and
University of California, San Diego}
\maketitle

\begin{abstract}
Generating diverse and natural human motion is one of the long-standing goals for creating intelligent characters in the animated world.
In this paper, we propose a self-supervised method for generating long-range, diverse and plausible behaviors to achieve a specific goal location. 
Our proposed method learns to model the motion of human by decomposing a long-range generation task in a hierarchical manner. 
Given the starting and ending states, a memory bank is used to retrieve motion references as source material for short-range clip generation. 
We first propose to explicitly disentangle the provided motion material into style and content counterparts via bi-linear transformation modelling, where diverse synthesis is achieved by free-form combination of these two components.
The short-range clips are then connected to form a long-range motion sequence.
Without ground truth annotation, we propose a parameterized bi-directional interpolation scheme to guarantee the physical validity and visual naturalness of generated results.
On large-scale skeleton dataset, we show that the proposed method is able to synthesise long-range, diverse and plausible motion, which is also generalizable to unseen motion data during testing. 
Moreover, we demonstrate the generated sequences are useful as subgoals for actual physical execution in the animated world. 
Please refer to our project page for more synthesised
results~\footnote{\href{https://sites.google.com/view/hsnms}{https://sites.google.com/view/hsnms}}.

\end{abstract}

\section{Introduction}
\label{sec:intro}
\begin{figure*}[t]
\centering 
\includegraphics[width=1.0\linewidth]{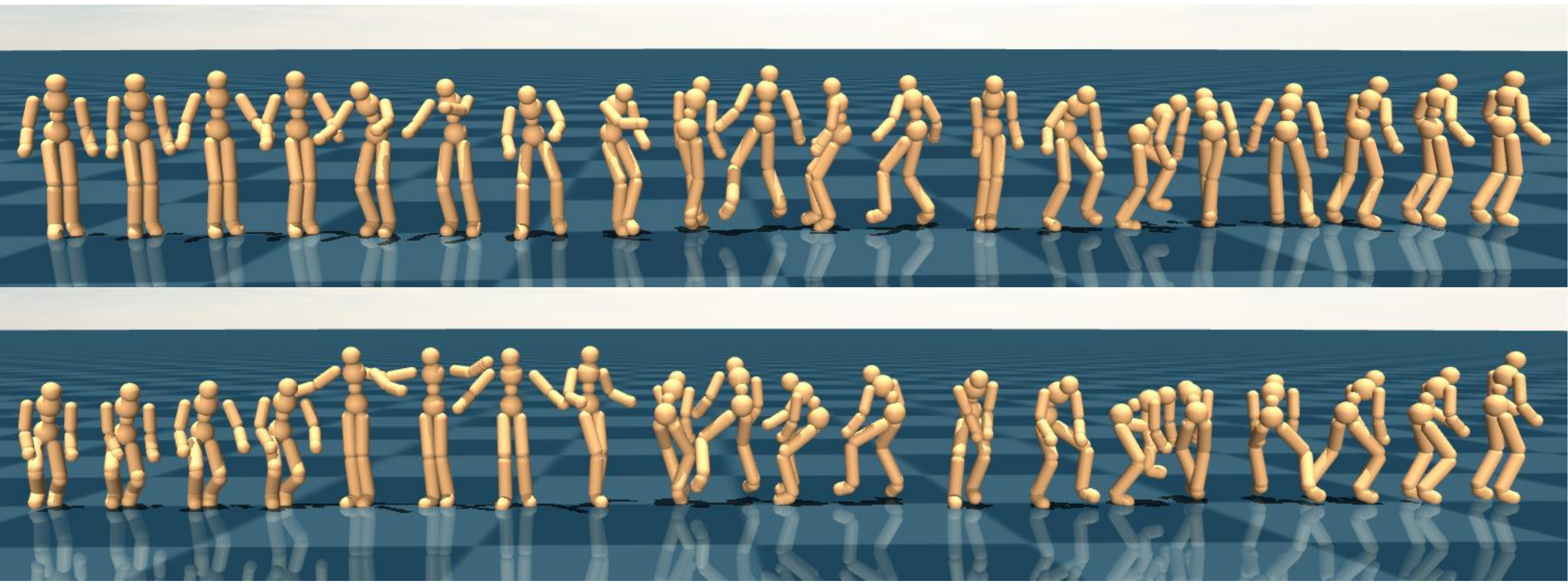}
\caption{Visualization of motion sequences generated by proposed hierarchical style-based networks. The generation is achieved by transferring the ``style'' of multiple subsequences to new ones and then connect these clips smoothly in a temporal sequential manner. We present two synthesized motion sequences, which consist of the same motion ``content'', but capture different motion ``style''. The specific definition of ``content'' and ``style'' is described in the method part. The long-range and diverse motion generation is thus achieved by free-form composition of ``content'' and ``style'' and connecting short-range clips to a long-range one.}
\label{fig:intro_fig}
\end{figure*}

Human motion is naturally continuous in time and diverse between different individuals. The same action performed by different people can look very different, and even performed by the same actor twice could hardly be identical (Fig.~\ref{fig:intro_fig}). Capturing this diverse and stylized motion has been a long-standing demand in animation production and video games~\cite{DBLP:journals/tog/HoldenKS17}. By generating this natural motion automatically, it provides useful tools for player customization of action skills in video games~\cite{DBLP:journals/tog/HoldenKS17,DBLP:journals/tog/ZhangSKS18}. However, synthesizing long-range and diverse motion is an extremely challenging task. On one hand, given the nature of multi-modal human behaviours, it is difficult to generate diverse motion without access to the distribution of motion states. On the other hand, it is common in sequential generative models that the error will accumulate through time, which restricts the maximum length of the generated motion sequence~\cite{DBLP:conf/nips/FinnGL16}.

To generate natural and diverse motion, researchers have proposed statistical models based on optimization~\cite{DBLP:journals/tog/XiaWCH15,DBLP:journals/tog/LiWS02,DBLP:conf/siggraph/BrandH00}. For example, Style Machine is a probabilistic generative model introduce in~\cite{DBLP:conf/siggraph/BrandH00}. It is capable of generating motion with different styles (e.g., motion of novice ballet or modern dance of an expert) by optimizing with a cross-entropy framework. Motivated by this work, Motion texture~\cite{DBLP:journals/tog/LiWS02} further proposed a two-level statistical model designed to capture diverse motion transitions, which achieves visually appealing state switching within seen sequence clips. However, these optimization-based methods are commonly restricted by the optimization complexity and can hardly be applied with large-scale dataset. Meanwhile, it is hard for these approaches to generalize to unseen distribution of data. 

To adapt and generalize to large-scale data, deep neural networks are utilized in several recent works~\cite{DBLP:journals/tog/HoldenSK16,DBLP:journals/tog/HoldenKS17,DBLP:journals/tog/ZhangSKS18}. For example, Holden et al.~\cite{DBLP:journals/tog/HoldenSK16} introduced to synthesize the motion sequence using deep networks by first taking the control signal as inputs, the outputs of the deep networks are then furthered edited via image-based style transfer techniques~\cite{DBLP:conf/cvpr/GatysEB16}. Although the adopted style transfer approach works well for transferring texture and color in the image domain, it does not necessarily generalize to motion stylization. Without carefully designed training strategies for post-processing, unnatural pose and other artifacts, e.g., foot sliding, is commonly observed during motion synthesis~\cite{DBLP:journals/cga/HoldenHKK17}.

In this work, we present hierarchical style-based networks, which leverage the large spectrum of human behaviours from unlabeled and unsegmented data to generate long-range, diverse and visually plausible motion sequences (as shown in Figure~\ref{fig:intro_fig}). Our framework is in a 2-level hierarchical structure: (i) Locally, our network first generates multiple short-range motion sequences independently; (ii) Globally, these short-range motion are then connected sequentially in time to a long-range motion sequence. 

For obtaining diverse long-range motion sequences, we will first need to generate diverse short-range motion clips. To achieve this, we propose to disentangle the feature representation for short-range motion to two parts, representing the content and the style of the motion. The \textit{content} refers to the specific action performed at each step, e.g., walking, running and dancing.
The \textit{style} refers to some pattern/property existing throughout the whole sequence, which keeps constant along with time.
Take the walking motion for example: senior individuals and child walking sequences demonstrate two different styles for the same ``walking'' action. With this disentangled representation, we can obtain diverse motion from the free-form combination of motion content vectors and style vectors. However, labeling content and style is expensive and most of the time it is even hard to define the style semantically. Thus, instead of defining the style, we propose a self-supervised learning approach to automatically discover the disentanglement between content and style. Specifically, we utilize a bi-linear transformation to explicitly decompose the motion feature into two components, and the two features are then combined together to reconstruct the input motion in an auto-encoder structure. The key for disentanglement is that we enforce the intermediate style feature to be consistent in time.

Given the synthesised short-range clips, we propose to connect them sequentially in time with motion interpolation between each two clips. Our network takes the starting and ending states from two different clips as inputs. To interpolate between these two states, we propose a parameterized representation which maintains the component scale of the generated sequences (e.g., bone lengths are fixed along time). Specifically, the representation is parameterized by a bi-directional LSTM model, which concurrently leverages the motion information of starting and ending states. In this way, we can obtain more plausible and visually appealing interpolation results.

Extensive experiments are conducted on large-scale human motion datasets~\cite{DBLP:journals/tog/HoldenSK16}.
We show that the proposed method outperforms existing motion generation baselines in terms of synthesis length, diversity and plausibility, e.g., being useful as sub-goals for actual physical execution in the animated world.
We also demonstrate the proposed model is capable of synthesising novel motion based on unseen data without additional fine-tuning procedure, which indicates the generalization ability of our method.

\section{Related Work}

\textbf{Motion Interpolation.}
Given start and end states, this task aims to synthesize intermediate states which smoothly translate between them~\cite{DBLP:conf/icml/UrtasunFGPDL08}.
For video interpolation~\cite{DBLP:conf/iccv/LiuYTLA17,DBLP:journals/corr/abs-1905-10240,DBLP:conf/cvpr/MeyerWZGS15,DBLP:conf/iccv/NiklausML17}, similar as prediction task~\cite{DBLP:conf/nips/XuNY18,DBLP:conf/cvpr/XuNLCY18}, where start and end states are two consecutive frames, the final result is expected to increase frame rate of original video to a higher value.
Previous researches often utilize phase dynamics~\cite{DBLP:conf/cvpr/MeyerWZGS15}, flow based feature~\cite{DBLP:conf/iccv/LiuYTLA17} and other motion information~\cite{DBLP:conf/iccv/NiklausML17,DBLP:conf/mm/YanXNZY17,Xu_2020_CVPR} to facilitate this task.
Our work is different from this branch of work because there exists large motion gap between start and end states in our settings.
Another branch of work is video completion~\cite{DBLP:conf/eccv/CaiBTT18,DBLP:journals/corr/abs-1905-10240,DBLP:journals/pami/WexlerSI07}.
It receives two \emph{nonconsecutive} frames as input and aims to fill the motion gap between start and end states. 
\cite{DBLP:conf/eccv/CaiBTT18} firstly attempts to solve this task and more specifically, propose to select out a rational path in the latent space with BFGS~\cite{DBLP:journals/siamsc/ByrdLNZ95} algorithm.
\cite{DBLP:journals/corr/abs-1905-10240} incorporates the 3D convolution layers and LSTM network into a unified model, which tries to automatically find the  optimal results for intermediate frames.
Despite much progress has been made in this filed, the high dimensional data (i.e., video frames) severely restricts video completion within \emph{simple} and \emph{seen} motion categories.
However, we do not limit the start and end states belonging to the same sequence.
Meanwhile, we expect the interpolated sequence as diverse as possible meanwhile with natural transition between synthesised states.
This has not been deeply addressed in previous motion completion works~\cite{DBLP:journals/tii/XiaSLH19,DBLP:conf/iclr/LeeSSHL19}. 
As a potential downstream application, our model could be used to construct motion planning~\cite{DBLP:journals/robotica/Myers83,DBLP:journals/corr/abs-2007-01738,DBLP:conf/ijcai/HuangXN18} algorithm.
Compared to goal-driven RL~\cite{DBLP:conf/nips/KulkarniNST16,DBLP:conf/iclr/LeeSSHL19}, our model gets rid of requirements hard to achieve, i.e., known dynamics of agent, which is more general and applicable to more motion planning scenarios.

\textbf{Motion Synthesis in Computer Graphics.} 
In the context of computer graphics, there is a branch of researches~\cite{DBLP:conf/sca/KovarG03,DBLP:conf/sca/ParkSS02,DBLP:journals/tog/LevineWHPK12,DBLP:journals/jvca/TanT12,DBLP:journals/tog/HoldenSK16} which also concentrate on motion generation, i.e., obtaining a continuous trajectory from a discrete set of poses.
Our work shares similar target with this branch of work.
However, we would like to emphasise that these works~\cite{DBLP:journals/jvca/TanT12,DBLP:conf/sca/KovarG03,DBLP:conf/sca/ParkSS02} are in parallel with ours and have completely different research routine on this task.
More specifically, graphics methods~\cite{DBLP:conf/siggraph/BrandH00,DBLP:journals/tog/LiWS02} focus on formulating an optimal and explicit statistic modelling framework, which is not easy scale-up to large datasets and unseen motion.
For example, \cite{DBLP:conf/siggraph/BrandH00} introduce a fully data-driven method for articulated motion generation, which needs online optimization if given new demonstration data as inputs. 
The involved learning procedure is relatively more complex compared to ours and needs careful hyperparameter tuning. 
On the contrary, our model combines both advantages of deep model and purely data-driven method.
Another related work~\cite{DBLP:journals/tog/LiWS02} proposes a statistical model for approximation of original motion distribution, which is represented by a transition matrix indicating the possibility of state change. 
This method requires all motion data are available for generation, which is in turn restricted to seen data. 
Differently, our model could generalize to unseen data, which implies more practical value for downstream application.

\section{Method}
We aim to synthesize visually natural, diverse and long-range motion sequence constrained by a pair of starting/ending states in a hierarchical manner.
We denote a motion sequence with length of $M$ as $\mathbf{S}=[\mathbf{s}_{1},...,\mathbf{s}_{i},...,\mathbf{s}_{M}]$, where $\mathbf{s}_{i}$ is the state at time stamp $i$.
The starting and ending states are denoted as $\mathbf{s}_{S}$ and $\mathbf{s}_{E}$ respectively.
Note that $\mathbf{s}\in \mathbb{R}^{J\times3}$, where $J$ refers to the number of joints, which is represented with the x-y-z Cartesian coordinate.

\begin{figure*}[t]
\centering 
\includegraphics[width=1.0\linewidth]{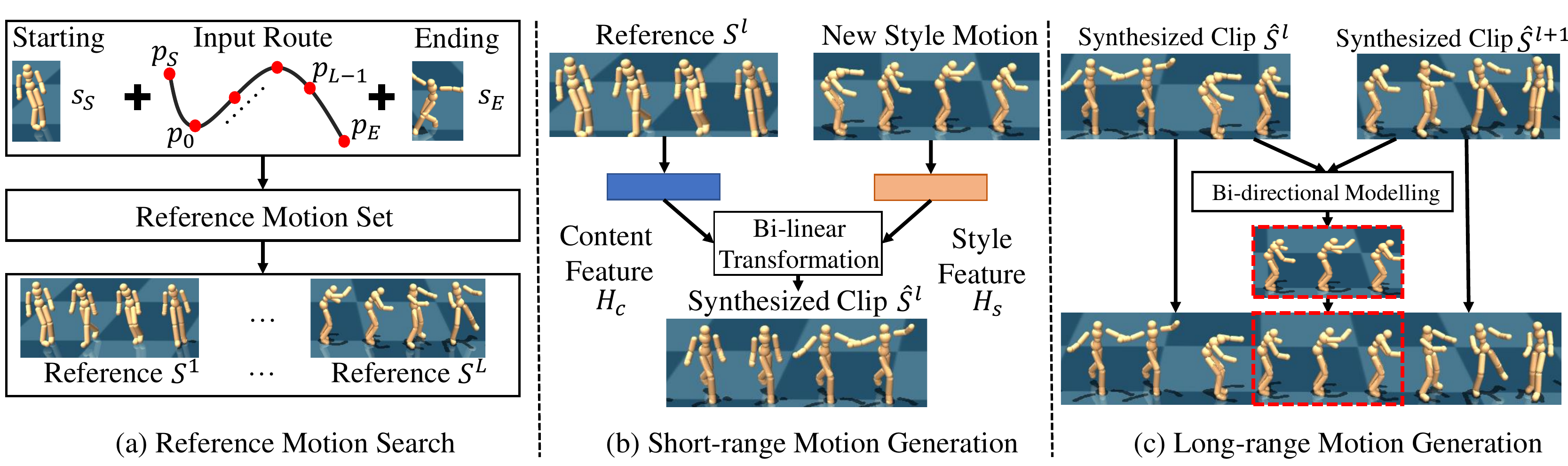}
\caption{\textbf{Hierarchical style-based motion synthesis framework.} (a) \emph{Reference Motion Search}: Given starting and ending states, we search for reference subsequences in the training dataset for generation; (b) \emph{Short-range Motion Generation}: given each reference subsequence, we generate a novel subsequence with motion style transfer; (c) \emph{Long-range Motion Generation}: all synthesized subsequences are connected together in time with bi-directional modelling.}
\label{fig:frame}
\end{figure*}

Our motion synthesis framework contains 3 steps as shown in Fig.~\ref{fig:frame}: (i) \emph{Reference Motion Search}. Given the input starting/ending states, we  first divide the route to $L$ segments by adding $L-1$ sub-goals in between on the ground; then, for a segment $l$, sampling 1 reference subsequence $\mathbf{S}^{l}\in \mathbb{R}^{M\times J\times 3}$ from the training dataset for motion generation, where $M$ represents the subsequence length.
Each sub-goal is represented by a spatial point $\mathbf{p}^{l}\in \mathbb{R}^{2},l=1,..,L-1$.  We denote $\mathbf{p}_{S}$ and $\mathbf{p}_{E}$ as the projected locations of root joint of $\mathbf{s}_{S}$ and $\mathbf{s}_{E}$.  (ii) \emph{Short-range Motion Generation}. Toward diverse motion generation, when generating each subsequence $l$, we advocate a synthesis paradigm in the motion style transfer manner, which keeps the content identical to that of $\mathbf{S}^{l}$ while changing the style based on another subsequence that is randomly sampled from the dataset. We denote the output short-range subsequence as $\hat{\mathbf{S}}^{l}$. (iii) \emph{Long-range Motion Generation.} Given the  short-range motion subsequences, we connect each two consecutive subsequences by adding a transitional motion in between. We denote the transitional motion between 
$\hat{\mathbf{S}}^{l}$ and $\hat{\mathbf{S}}^{l+1}$ as $\hat{\mathbf{S}}^{l,l+1}\in \mathbb{R}^{N\times J\times 3}$, where $N$ is the length of the transitional sequence. The final long-range generated sequence is presented as: $[\mathbf{s}_{S},\hat{\mathbf{S}}^{0,1},\hat{\mathbf{S}}^{1},...,\hat{\mathbf{S}^{l}},\hat{\mathbf{S}}^{l,l+1},\hat{\mathbf{S}}^{l+1},...,\hat{\mathbf{S}}^{L},\hat{\mathbf{S}}^{L,L+1},\mathbf{s}_{E}]$.

\textbf{Reference Motion Search.}
The searching procedure is conducted as follows:
(i) \emph{Reference length calculation}: For a reference subsequence $\mathbf{S}$ in the dataset, we calculate the distance (denoted as $d$) between the first and last states of $\mathbf{S}$. The minimum and maximum values are denoted as $d_{min}$ and $d_{max}$ respectively. 
(ii) \emph{Sub-goal sampling}: We randomly sample $L-1$ sub-goals on the ground, $[\mathbf{p}_{1},...,\mathbf{p}_{L-1}]$. The distance between two consecutive sub-goals (denoted as $d_{l}$) is sampled from $\mathcal{U}(d_{min}, d_{max})$. $\mathcal{U}()$ is the uniform distribution. The direction specified by the vector, $\overrightarrow{\mathbf{p}_{l}\mathbf{p}_{l+1}}=\mathbf{p}_{l+1}-\mathbf{p}_{l}$, is sampled from $\mathcal{U}(-\frac{\pi}{2},\frac{\pi}{2})$. $\mathbf{
s}_{S}$ and $\mathbf{s}_{E}$ are translated with an offset respectively, to make sure that the length of $\overrightarrow{\mathbf{p}_{S}\mathbf{p}_{1}}$ and $\overrightarrow{\mathbf{p}_{L-1}\mathbf{p}_{E}}$ (denoted as $d_{S}$ and $d_{E}$) satisfy $d_{min}<d_{S},d_{E}<d_{max}$.
(iii) \emph{Subsequence match}: For each sub-goal pair ($\mathbf{p}_{l},\mathbf{p}_{l+1}$), we select the reference subsequence whose length matches $d_{l}$ within a tolerance $\sigma$, i.e., $d_{l} - \sigma < d < d_{l}$ and $\sigma=0.05$. 
Finally, the selected subsequence $\mathbf{S}^{l}$ is rotated to match  $\overrightarrow{\mathbf{p}_{l}\mathbf{p}_{l+1}}$.

\subsection{Short-range Motion Generation}
In this section, we present the details of generating short-range motion clips.
This part is formulated as a motion style transfer task, i.e., new motion clips are synthesised via altering the style of reference subsequence $\mathbf{S}^{l}$ while keeping their motion contents unchanged. 
Recall that the \textit{content} refers to the specific action performed at each step, e.g., walking, running and dancing, while the \textit{style} refers to some pattern/property existing throughout the whole sequence, which keeps constant along with time.
The free-form combination of content and style information is used for diverse synthesis of subsequences. We denote the generated subsequence as  $\hat{\mathbf{S}}^{l}=\{\hat{\mathbf{s}}_{i}^{l}\}_{i=1}^{M}$. 
Without annotation of style, we first propose to learn corresponding representations in a disentangling manner.

\begin{figure*}[t]
\centering 
\includegraphics[width=1.0\linewidth]{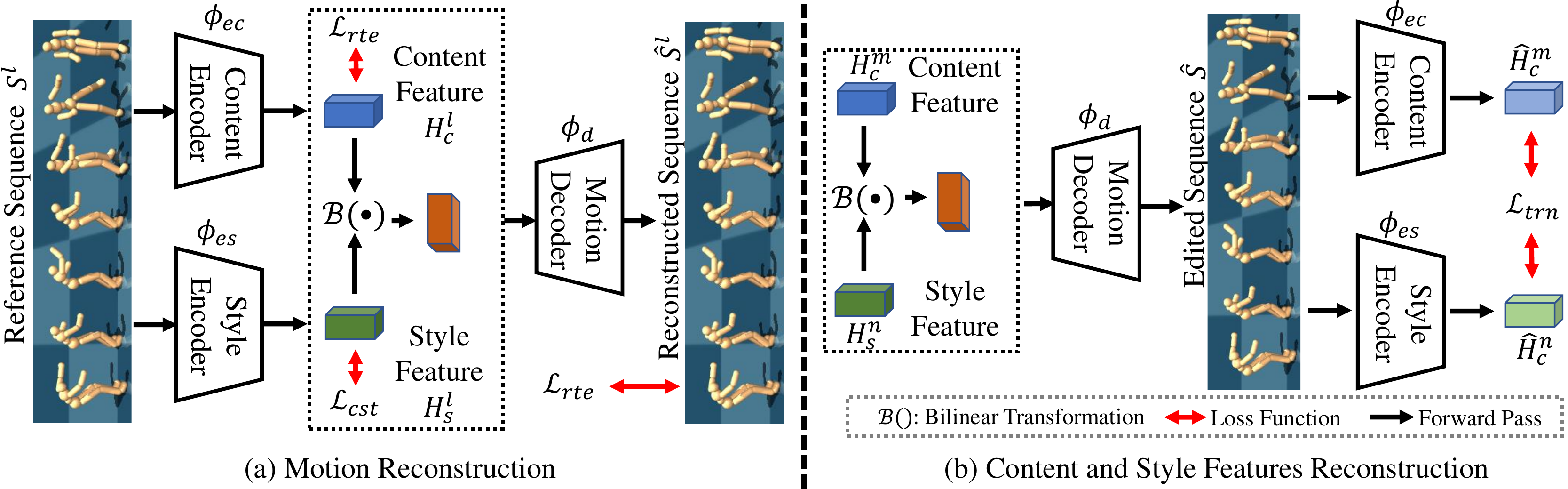}
\caption{\textbf{Short-range motion generation.} (a) We train an auto-encoder for motion sequence $\mathbf{S}^{l}$ reconstruction. The bi-linear transformation is utilized for content and style disentanglement. (b) Training with content and style features from different sources of motion subsequences. Given two features as inputs, we generate the motion which can be used to reconstruct the feature inputs. The training objective is defined on the content and style feature reconstruction error $\mathcal{L}_{trn}$.}
\label{fig:short}
\end{figure*}

\textbf{Content and Style Disentanglement.}
\textcolor{black}{As shown in Fig.~\ref{fig:short} (a), two encoders, $\phi_{ec}$ and $\phi_{es}$, are used to extract the content and style features as follows,}
\begin{equation}
\mathbf{H}_{c}^{l}=\phi_{ec}(\mathbf{S}^{l}),\mathbf{H}_{s}^{l}=\phi_{es}(\mathbf{S}^{l}),
\end{equation}
\textcolor{black}{where $\mathbf{S}^{l}$ is an input subsequence, $\mathbf{H}_{c}^{l}=\{\mathbf{h}_{c,i}^{l}\}_{i=1}^{M}$ and $\mathbf{H}_{s}^{l}=\{\mathbf{h}_{s,i}^{l}\}_{i=1}^{M}$ are the content and style features respectively, $M$ represents the number of steps in each subsequence.}
Inspired by the success of content and style separation in character and image~\cite{DBLP:conf/nips/TenenbaumF96} fields, we propose to reconstruct the motion with a bi-linear transformation scheme, where each time step $i$ is represented as:

\begin{equation}
\hat{\mathbf{s}}_{i}^{l}=\phi_{d}(\mathcal{B}(\mathbf{h}_{c,i}^{l},\mathbf{h}_{s,i}^{l})),
\end{equation}
where $\mathcal{B}(\cdot)$ is the bi-linear transformation~\cite{DBLP:conf/nips/TenenbaumF96}, i.e., $\mathcal{B}(\mathbf{h}_{c,i}^{l},\mathbf{h}_{s,i}^{l})=\mathbf{h}_{c,i}^{l}\mathcal{W}(\mathbf{h}_{s,i}^{l})^\mathsf{T}$ and 
$\mathcal{W}\in \mathbb{R}^{C_{ctn}\times C_{out}\times C_{sty}}$ is a bi-linear transformation weight.
$\phi_{d}$ is the motion decoder. 
As a two-factor method, the bi-linear transformation possesses an elegant mathematical property, i.e., separability: their outputs are linear in either component when the other is kept unchanged.
Facilitated by bi-linear transformation, the contribution of two components can be effectively disentangled and fused into a representative latent feature that is generalizable to unseen factor modes with new contents.

\textbf{Training Objective.} Encoder and decoder are jointly trained by a L2 reconstruction loss, i.e., $\mathcal{L}_{rec}=||\mathbf{S}^{l}-\hat{\mathbf{S}}^{l}||_{2}^{2}$.
To guarantee the style consistency of the reconstructed subsequence, an explicit L2 loss is incorporated into training procedure, i.e., \textcolor{black}{$\mathcal{L}_{cst}=||\mathbf{\hat{h}}_{s,i}^{l}-\mathbf{\hat{h}}_{s,j}^{l}||_{2}^{2}$, where $i,j$ are randomly sampled two style features corresponding to time step $i$ and $j$ respectively.}
Meanwhile, the moving route $\mathbf{c}^{l}$ (including root position and velocity, foot position and velocity~\cite{DBLP:journals/tog/HoldenSK16}) is extracted as a part of content feature and trained with a L2 loss to prevent foot sliding, i.e., $\mathcal{L}_{rte}=||\mathbf{c}^{l}-\hat{\mathbf{c}}^{l}||_{2}^{2}$, where $\mathbf{c}^{l}$ is the input control signal.
\textcolor{black}{Note that the moving route is treated as a part of content feature $\mathbf{h}_{c}^{l}$, i.e., $\mathbf{c}^{l}\in \mathbf{h}_{c}^{l}$.} The training pipeline of this motion auto-encoder is shown in Fig.~\ref{fig:short}(a).

At each iteration we sample a new batch of data for training, to further guarantee the consistency at feature level.
Suppose a subsequence (denoted as $\hat{\mathbf{S}}$) is generated with the content $\mathbf{H}_{c}^{m}$ of $\mathbf{S}^{m}$ and style $\mathbf{H}_{s}^{n}$ of $\mathbf{S}^{n}$.
$\hat{\mathbf{S}}$ is subsequently fed into $\phi_{ec}$ and $\phi_{es}$ to obtain reconstructed features, i.e., $\hat{\mathbf{H}}_{c}^{m}$ and $\hat{\mathbf{H}}_{s}^{n}$, as shown in Fig.~\ref{fig:short}(b).
This part is trained with
$\mathcal{L}_{trn}=||\mathbf{H}_{c}^{m}-\hat{\mathbf{H}}_{c}^{m}||_{2}^{2}+||\mathbf{H}_{s}^{n}(\mathbf{H}_{s}^{n})^{\mathsf{T}}-\hat{\mathbf{H}}_{s}^{n}(\hat{\mathbf{H}}_{s}^{n})^{\mathsf{T}}||_{2}^{2}$.
The latter part for style consistency is inspired from Gram Matrix utilized in image translation~\cite{DBLP:conf/cvpr/GatysEB16}. 
The high-order statistics of style feature has been proven to be more critical for translation in image domain. The final training objective is presented as follows,
\begin{equation}
\mathcal{L} = \mathcal{L}_{rec}+0.01\mathcal{L}_{cst}+0.5\mathcal{L}_{rte}+\mathcal{L}_{trn}.
\label{Eqn:short_loss}
\end{equation}
Note that both the encoder and decoder are jointly trained under the supervision of $\mathcal{L}_{trn}$.
We present the model architecture details in the supplementary material.

\begin{figure*}[t]
\centering 
\includegraphics[width=1.0\linewidth]{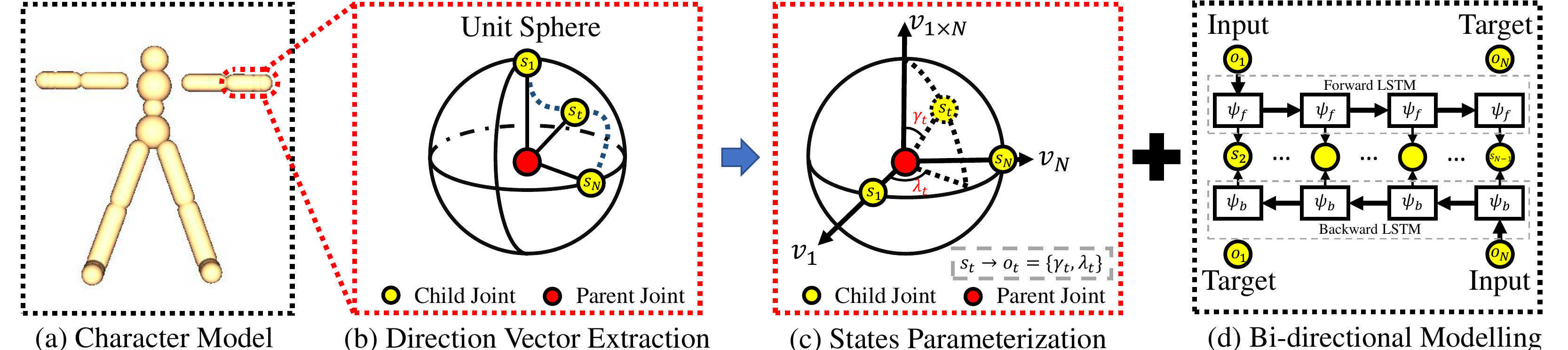}
\caption{\textbf{Proposed model for long-range motion generation.} Starting with skeleton model of human subject (leftmost), we extract the corresponding direction vector from 3D coordinate (second part), which are parameterized by the starting/ending states of one subsequence (third part). This leads to more compact solution space with learnable parameters ($\gamma,\lambda$), which are generated by the proposed bi-directional model.}
\label{fig:inter}
\end{figure*}

\subsection{Long-range Motion Generation}
This part refers to connecting edited short-range subsequences ($\hat{\mathbf{S}}^{l}$, $\hat{\mathbf{S}}^{l+1}$) into a long-range one with interpolated subsequence $\hat{\mathbf{S}}^{l,l+1}$.
To achieve smooth and natural transition, we propose to Parameterize the original 3D coordinate space as shown in Fig.~\ref{fig:inter}, which leads to a more compact output space.

\textbf{Parameterized Representation.} 
Given the state $\mathbf{s}_{t}$ at time step $t$ we first obtain the direction vector $\mathbf{v}_{t}(p,q)=\frac{\mathbf{s}_{t}(p)-\mathbf{s}_{t}(q)}{||\mathbf{s}_{t}(p)-\mathbf{s}_{t}(q)||_{2}^{2}}$, where $(p,q)$ corresponds to one child-parent pair of character joint according to the skeleton topology (Fig.~\ref{fig:inter}(b)). 
\textcolor{black}{The transition subsequence is generated in the direction vector space.}
Given two vectors ($\mathbf{v}_{1},\mathbf{v}_{N}$) as bases (the length of transition is assumed to be $N$), interpolation is essentially the combination of $\mathbf{v}_{1}$ and $\mathbf{v}_{N}$.
The interpolation procedure is inspired from Quaternion Slerp~\cite{DBLP:conf/siggraph/Shoemake85} but generalized to non-linear situation.
As shown in Fig.~\ref{fig:inter}(c), supposing $\mathbf{v}_{1}$ and $\mathbf{v}_{N}$ are two bases in 3D space, the third basis is obtained by outer product: $\mathbf{v}_{1\times N}=\mathbf{v}_{1}\times \mathbf{v}_{N}$. An arbitrary direction vector $\mathbf{v}_{t}$ could be represented in the parameterization manner, 
\begin{equation}
\begin{small}
\mathbf{v}_{t}=\frac{\sin(1-\gamma_{t})\Lambda}{\sin\Lambda}(\frac{\sin(1-\lambda_{t})\Omega}{\sin\Omega}\mathbf{v}_{1}+\frac{\sin(\lambda_{t}\Omega)}{\sin\Omega}\mathbf{v}_{N})+\frac{\sin(\gamma_{t}\Lambda)}{\sin\Lambda}\mathbf{v}_{1\times N},
\label{Eqn:non-linear}
\end{small}
\end{equation}
where $\Omega$ is directed angle between $\mathbf{v}_{1}$ and $\mathbf{v}_{N}$, $\Lambda=\pi/2$ because of outer product.
$\mathbf{v}_{t}$ is thus parameterized by $\mathbf{v}_{1}$ and $\mathbf{v}_{N}$ with $(\gamma_{t}, \lambda_{t})$.
Note that $\gamma_{t}, \lambda_{t}$ are two learnable parameters in our work, which is modelled by a bi-directional LSTM~\cite{DBLP:journals/neco/HochreiterS97} introduced in following paragraph.

\textcolor{black}{Intuitively, the interpolation defined by Eqn.~\ref{Eqn:non-linear} is analogous to locating on the earth with the longitude and latitude. 
The Eqn.~\ref{Eqn:non-linear} conducts quaternion slerp twice, where the first one, inside the brackets, decides the ``longitude'' ($\lambda$) and the second one, outside the brackets, decides the ``latitude'' ($\gamma$).}
One one hand, the output dimension at each time stamp is reduced from 4J (quaternion) to 2J ($\gamma,\lambda$), leading to a more compact solution space. 
On the other hand, outputs are naturally valid, i.e., unit vector required by direction vector, which avoids additional normalization procedure. 

\textbf{Bi-directional Modelling.}
As shown in Fig.~\ref{fig:inter}(d), we utilize two LSTM~\cite{DBLP:journals/neco/HochreiterS97} networks (forward and backward) to achieve natural and plausible interpolation in the parametrization space ($\mathbf{o}=\{\lambda,\gamma\}$).
The forward one (denoted as $\psi_{f}$) predicts future states with $\mathbf{o}_{1}$ as origin and $\mathbf{o}_{N}$ as target.
The counterpart (denoted as $\psi_{b}$) takes the reverse direction, i.e., $\mathbf{o}_{N}$ as origin and $\mathbf{o}_{1}$ as target.
The whole modelling procedure is executed as follows,
\begin{equation}
\hat{\mathbf{o}}_{t+1,f}=\psi_{f}(\hat{\mathbf{o}}_{t,f}, \mathbf{o}_{N}),\hat{\mathbf{o}}_{t-1,b}=\psi_{b}(\hat{\mathbf{o}}_{t,b}, \mathbf{o}_{1}).
\label{eq:bi-pred}
\end{equation}
The complementary pair of forward/backward outputs, i.e., $(\mathbf{\hat{o}}_{t,f},\mathbf{\hat{o}}_{N-t,b})$ are firstly converted back to the x-y-z coordinate space $(\mathbf{\hat{s}}_{t,f},\mathbf{\hat{s}}_{N-t,b}$) with forward kinematics~\cite{QuaterNet}, and then fused as follows, 
\begin{equation}
\mathbf{\hat{s}}_{t}=\psi_{fse}(\mathbf{\hat{s}}_{t,f},\mathbf{\hat{s}}_{N-t,b}),
\end{equation}
\begin{equation}
\mathcal{L}_{pos}=||\mathbf{s}_{t}-\hat{\mathbf{s}}_{t}||_{2}^{2}.
\end{equation}
With supervision signal in the position space, the topology information of character skeleton is better utilized and motion generation artifacts, e.g, foot sliding, could be explicitly punished during training procedure. 

\textbf{Adversarial Training: Generalizing beyond Training Sequence.} Above interpolation is trained with starting/ending pair belonging to the same sequence, i.e., with ground truth. However, proposed model should be tested with arbitrary state pairs for practical usage. It lacks ground truth signal for training under such condition.
We thus utilize adversarial training to facilitate generalization ability of motion interpolation.
More specifically, with $\mathbf{s}_{1},\mathbf{s}_{N}$ sampled from different sequences, $\psi_{f,b}$ give prediction of intermediate states as described above, which are regarded as fake sequences.
A motion discriminator $D$ is further proposed for adversarial training as follows,
\begin{equation}
\mathcal{L}_{D}=\frac{1}{2}((1-D(\mathbf{S}))^{2}+D^{2}(G(\mathbf{s}_{1},\mathbf{s}_{N}))),
\end{equation}
\begin{equation}
\mathcal{L}_{G}=(1-D(G(\mathbf{s}_{1},\mathbf{s}_{N})))^{2},
\end{equation}
where $G$ refers to proposed interpolation model in this section.
The discriminator architecture follows the work of Chen et.al.~\cite{DBLP:conf/cvpr/ChenTADMSR19}, i.e., built based on residual block and designed for synthesising more realistic 3D poses.

\paragraph{Implementation Details.} We train the models for short-range and long-range motion generation in a 2-step procedure. For the long-range motion generation, the adversarial training scheme is alternating the supervision between $\mathcal{L}_{pos}+0.01\mathcal{L}_{G}$ and $0.01\mathcal{L}_{D}$.
We adopt Adam~\cite{DBLP:journals/corr/KingmaB14} as the optimizer, where learning rate, learning decay and weight decay are set to $5e^{-4}, 0.97, 1e^{-5}$ respectively. Both short-range and long-range models are trained with 200 epochs.

\section{Experiments}
\label{sec:exp}

\begin{figure}[t]
\centering
\includegraphics[width=1\linewidth]{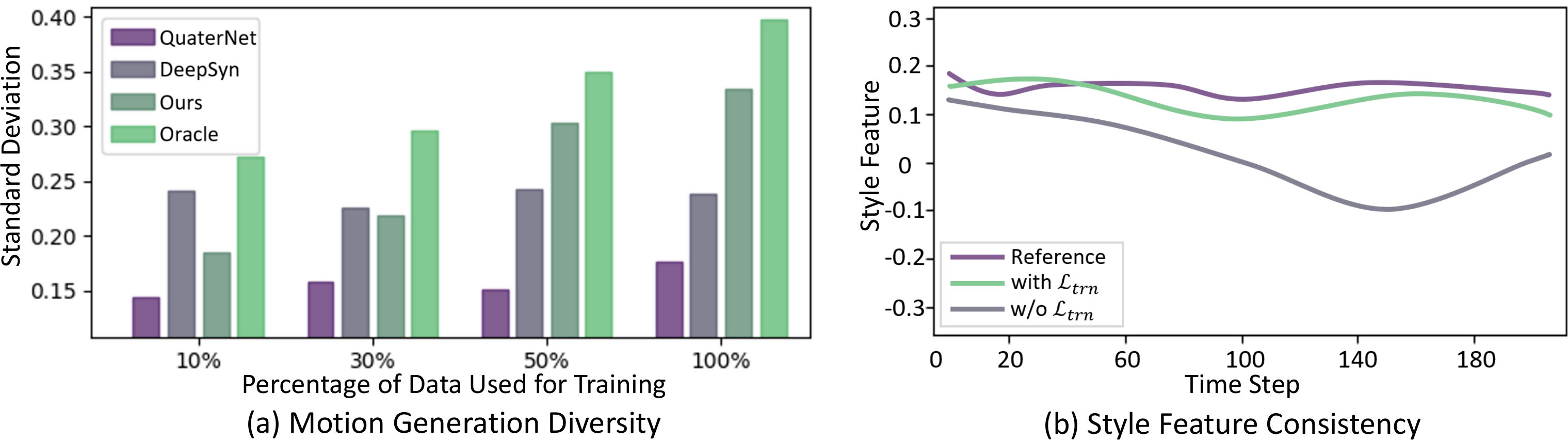}
\caption{\textbf{Short-range motion evaluation}: (a) Motion diversity evaluation. X-axis is the proportion of data used for training. Y-axis is the deviation of joint coordinates. Our model is able to synthesize more diverse sequences than baseline models.
(b) Style feature visualization. We can see that facilitated by the loss $\mathcal{L}_{trn}$, the style feature of synthesized sequence is closer to the input one (reference) and kept relatively constant.}
\label{fig:sfv}
\end{figure}

\subsection{Evaluation Settings}
\label{Sec:es}
\textbf{Datasets.}
We use the CMU Mocap dataset~\footnote{http://mocap.cs.cmu.edu/} and follow the processing procedure of Holden et.al.~\cite{DBLP:journals/tog/HoldenSK16}, where $M=120$ and $N=40$.
The dimension of state at a single time step $\mathbf{s}_{t}$ is 63, where $J=21$ with x-y-z Cartesian coordinates.
To demonstrate that our model could generate novel behaviour never seen during training, we keep a held-out reference set (denoted as $\mathcal{D}_{R}$) from training data (denoted as $\mathcal{D}_{T}$) for further testing.

\textbf{Baselines.} 
Considering the hierarchical structure of our model, we conduct comparison experiments at both levels, i.e., short-range and long-range.
For short-range generation, we compare our model with two strong baselines: the work of Holden et.al.~\cite{DBLP:journals/tog/HoldenSK16} (denoted as DeepSyn in this paper) and QuaterNet~\cite{QuaterNet}. 
Both baselines are retrained with the same data used in this work.
The input dimension is adjusted to match our data.
For long-range generation, we compare our model with temporal prediction baselines: HP-GAN~\cite{DBLP:conf/cvpr/BarsoumKL18} and MT-VAE~\cite{DBLP:conf/eccv/YanRVSSHYL18}.
For these two baselines, the ending states are feed as input with the default hyper-parameter setting.

\subsection{Evaluation of Short-range Generation}

\textbf{Generation Diversity Evaluation.} 
We compare our model with DeepSyn~\cite{DBLP:journals/tog/HoldenSK16} and QuaterNet~\cite{QuaterNet} to evaluate the motion diversity, under different percentages of data used for training.
As illustrated in Fig.~\ref{fig:sfv}(a), $10\%$, $30\%$, $50\%$ and $100\%$ training data are used respectively.
We calculate the averaged standard deviation of all joints where higher value indicates higher diversity.
QuaterNet~\cite{QuaterNet} achieves the lowest diversity under all settings.
DeepSyn~\cite{DBLP:journals/tog/HoldenSK16} keeps relative constant motion diversity which is comparable with $10\%$ training data.
On the contrary, the diversity of motion synthesized by our model constantly increases if more data used for training, which is mainly facilitated by the style-based synthesis model.

\begin{figure*}[t]
\centering 
\includegraphics[width=1.0\linewidth]{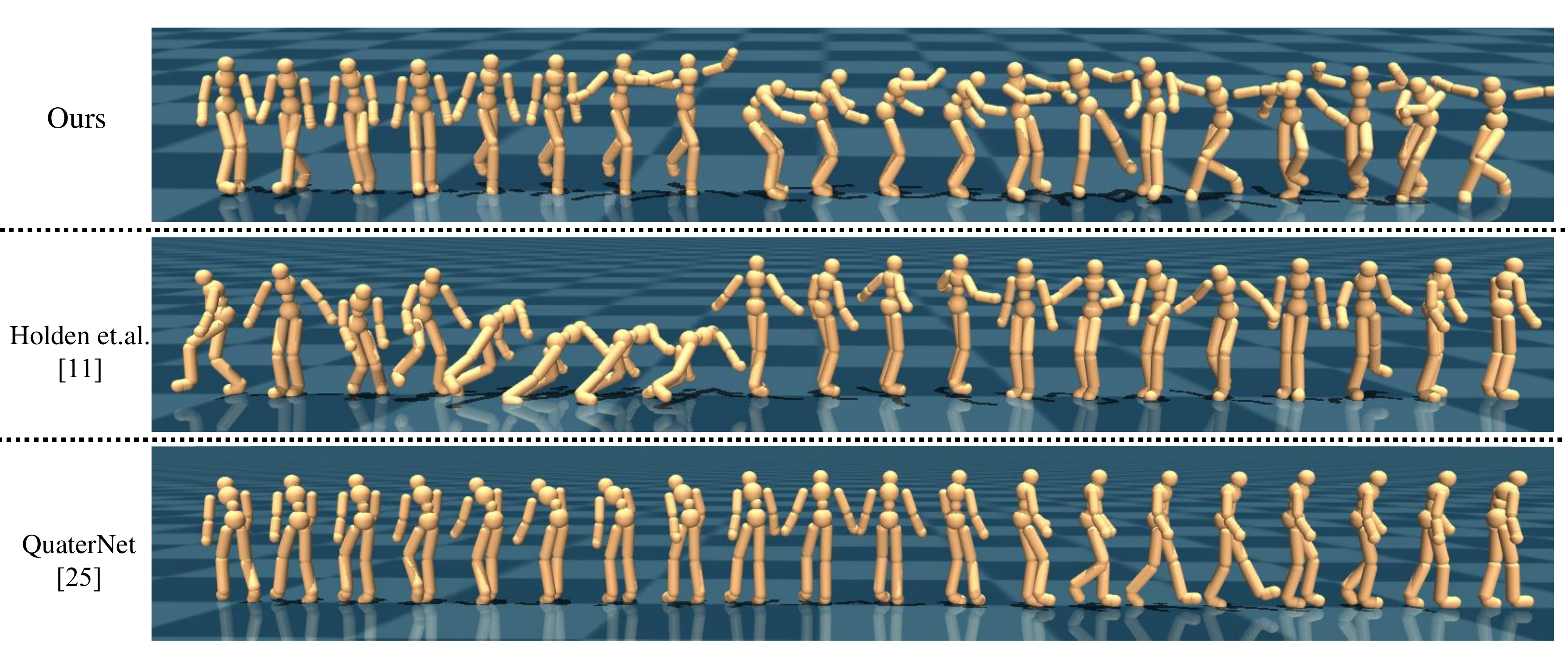}
\caption{Comparison of motion naturalness with DeepSyn~\cite{DBLP:journals/tog/HoldenSK16} and QuaterNet~\cite{QuaterNet}.}
\label{fig:lcoal}
\end{figure*}

\begin{figure*}[t]
\centering 
\includegraphics[width=1.0\linewidth]{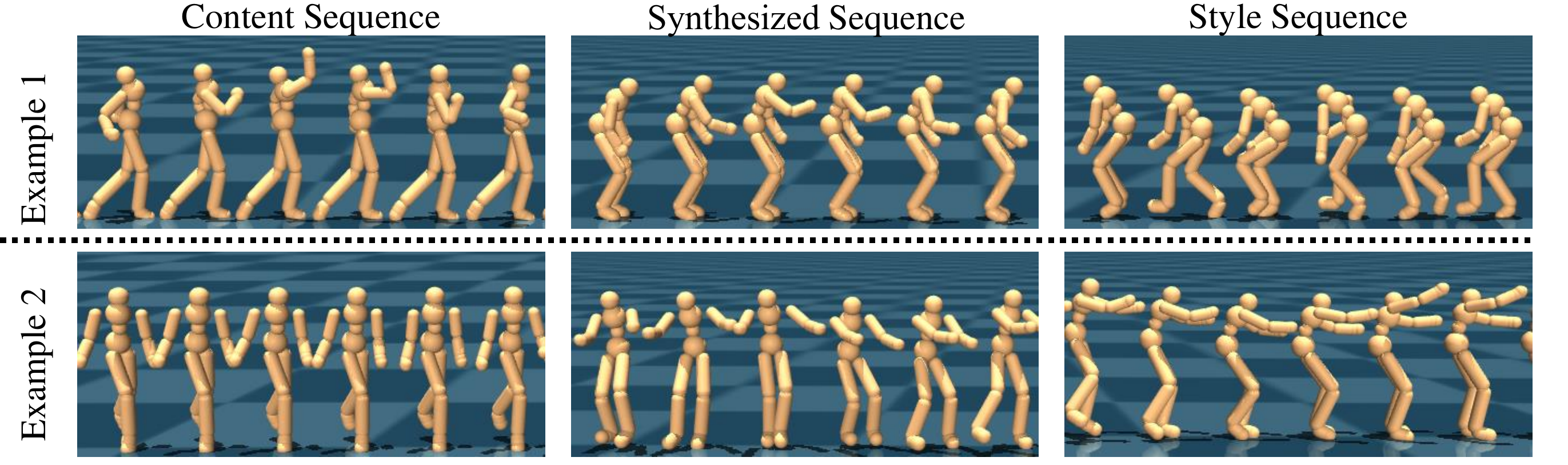}
\caption{\textbf{Visualization of generated short-range subsequences.} The synthesized short-range sequence (middle column) well preserves the detailed motion from content sequence (e.g., the arm moving in the first row) and constant pattern from style sequence (e.g., the arm lifting in the second row).}
\label{fig:style}
\end{figure*}

\textbf{Short-range Generation Visualization.}
Fig.~\ref{fig:lcoal} shows the short-range results (from top to bottom: our model, DeepSyn~\cite{DBLP:journals/tog/HoldenSK16} and QuaterNet~\cite{QuaterNet}).
We can observe that DeepSyn~\cite{DBLP:journals/tog/HoldenSK16} synthesizes an abnormal walking sequence with unnatural behavior (middle row in Fig.~\ref{fig:lcoal}, fall-down pose during synthesis).
QuaterNet~\cite{QuaterNet} is able to generate a visually natural walking sequence but struggles to produce diverse motion behaviour (last row in Fig.~\ref{fig:lcoal}, restricted to simple locomotion synthesis).
Different from all these models, our hierarchical style-based model achieves smooth transition throughout the whole sequence (first row in Fig.~\ref{fig:lcoal}), and provides natural and diverse motion (i.e., walking-turning-dancing) behaviours during generation.

\begin{figure*}[t]
\centering
\includegraphics[width=1\linewidth]{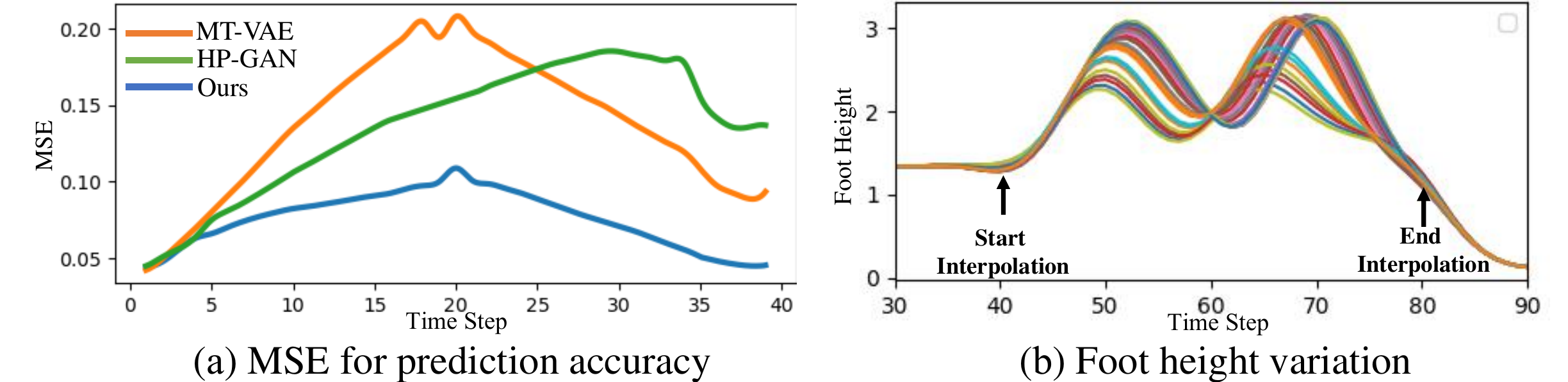}
\caption{Evaluation of long-range motion generation in terms of MSE (a) and foot height variation (b). (a) shows that our model (blue line) achieves lower interpolation error than baseline models. (b) demonstrates the non-linearity of interpolation results.}
\label{fig:fl}
\end{figure*}

\textbf{Style Feature Consistency Evaluation.}
As shown in Fig.~\ref{fig:sfv}(b), we plot one representative dimension of the learned style feature from a complete sequence.
The purple/green/gray curves refer to the style feature of reference sequence, synthesized sequence trained with and without $\mathcal{L}_{trn}$ respectively.
We can observe that training without the supervision of $\mathcal{L}_{trn}$ leads to the transferred style feature (gray curve) drifts away from the reference one (purple curve).
On the contrary, training with the supervision of $\mathcal{L}_{trn}$ (green curve) effectively facilitates transferring the style pattern from reference sequence to the synthesized one, i.e., the green curve is close to the purple curve.

\textbf{Style-based Generation Evaluation.}
As shown in Fig.~\ref{fig:style}, we provide two synthesized examples (middle part) which possess the general motion style from one subsequence (right part), and detailed motion pattern from another one (left part).
For the first sequence (top row in Fig.~\ref{fig:style}), the target style motion shows a walking sequence with back bent over (style), while the content motion (sampled from held-out set $\mathcal{D}_{R}$) is a sequence with arm waving.
We can notice that the synthesized motion (second column, top row) preserves the arm motion and learns the style pattern (back bent down) successfully.
For the second example (bottom row in Fig.~\ref{fig:style}), the target style motion shows a walking sequence with arm lifted horizontally (style), while the reference motion content (sampled from training set $\mathcal{D}_{T}$) is a regular walking sequence.
The synthesized motion (second column, bottom row) fully captures the style of upper body meanwhile preserves the walking motion from reference sequence.

\subsection{Evaluation of Long-range Motion Generation}

\begin{figure*}[t]
\centering 
\includegraphics[width=1.0\linewidth]{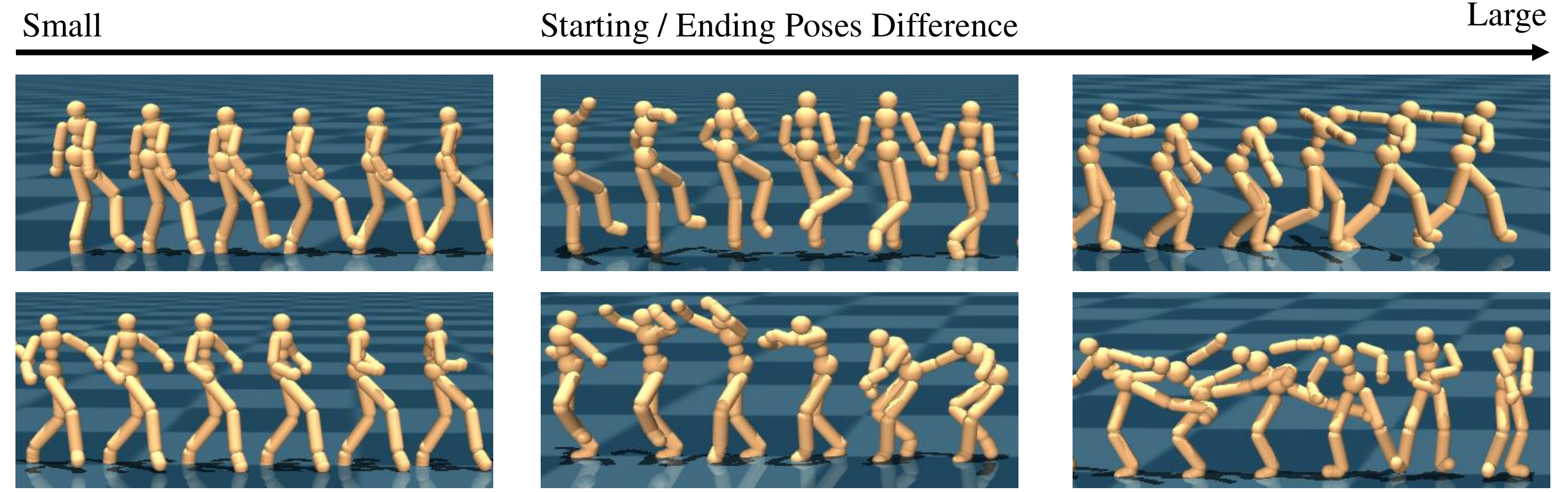}
\caption{\textbf{States transition visualization for evaluation of smoothness.} From left to right, the pose difference between starting and ending states becomes larger. Our model is able to generate smooth and natural transition under all situations.}
\label{fig:st}
\end{figure*}

\textbf{Interpolation Accuracy Evaluation.} 
We compare with two goal-conditioned prediction models (HP-GAN~\cite{DBLP:conf/cvpr/BarsoumKL18} and MT-VAE~\cite{DBLP:conf/eccv/YanRVSSHYL18}) evaluated by prediction accuracy.
The training data is a subset chosen from original one, consisting of locomotion, punching, kicking and dancing sequences.
The test data belongs to the same motion category of no overlap with training data.
We calculate the MSE value (the lower the better) for evaluation.
As shown in Fig.~\ref{fig:fl} (A), we can see that facilitated by the bi-directional modelling, our model could equally consider the contribution of starting/ending states, which leads to high transition accuracy (i.e., low MSE value) near the starting/ending states.
Facilitated by the proposed parameterization method (i.e., more compact output space), the interpolation accuracy outperforms both HP-GAN~\cite{DBLP:conf/cvpr/BarsoumKL18} and MT-VAE~\cite{DBLP:conf/eccv/YanRVSSHYL18} by a large margin.  

\textbf{Interpolation Sequence Visualization.}
Fig.~\ref{fig:st} demonstrates the interpolation results where starting and ending states are from different reference subsequences respectively.
The starting and ending states are sampled from held-out set ($\mathcal{D}_{R}$) for evaluation.
From left to right in Fig.~\ref{fig:st}, the pose difference between the starting and ending states gradually increases.
We can observe that our model is able to generate smooth and natural transition when starting and ending states are similar (left part in Fig.~\ref{fig:st}).
Moreover, when encountered large motion difference (right part in Fig.~\ref{fig:st}), e.g., from walking to dancing, turning back with a relatively large degree, our model still makes it to generate visually natural transition sequence.

\begin{figure*}[t]
\centering 
\includegraphics[width=1.0\linewidth]{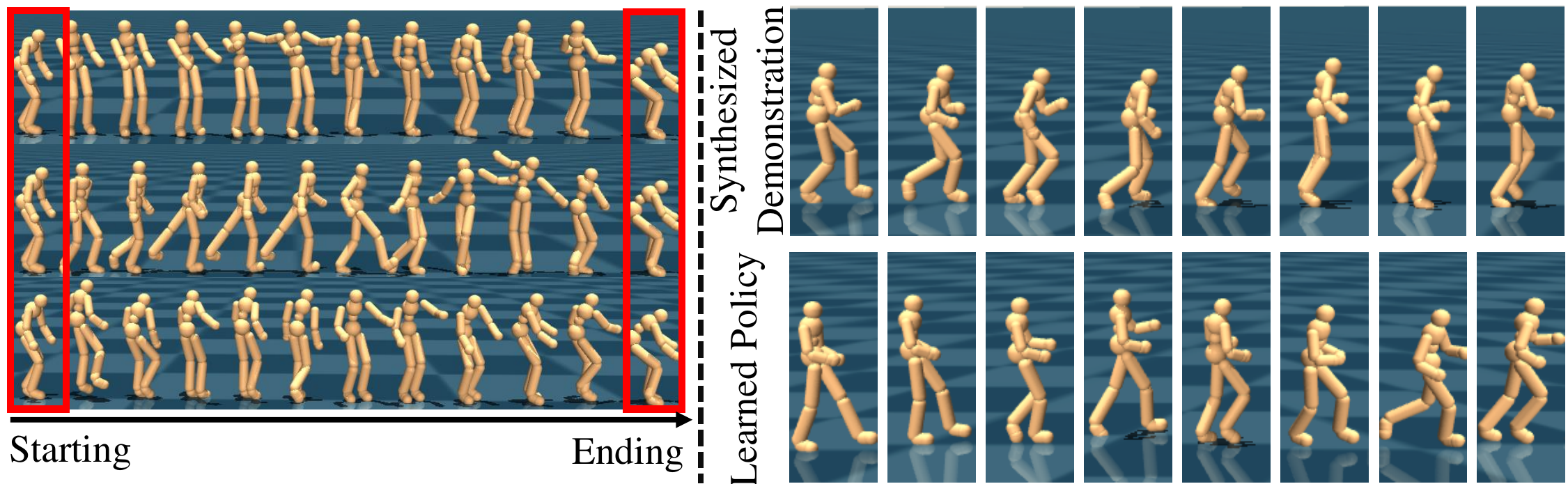}
\caption{Left: Diverse motion generation given the same starting and ending states. Right: Expert demonstration guidance for imitation learning.}
\label{fig:div}
\end{figure*}

\textbf{Non-linear Interpolation Verification.}
To evaluate whether our model learns non-linear interpolation between two sequences, we report the height variation of the right foot in a interpolation motion sequence.
Meanwhile, we randomly rotate the second sequence to show that our model is robust to a wide range of direction difference between two sequences.
As shown in Fig.~\ref{fig:fl}(B), we present multiple curves which correspond to different rotation angles.
All recorded curves are highly non-linear but smoothly changed between starting and ending points.
Moreover, our model adaptively changes foot height with different rotations, which leads to visually appealing interpolation results.

\textbf{Visualization of Final Synthesized Sequences.}
With both short-range and long-range motion generation, we are able to synthesize final sequences.
Recall that our model is constrained by given starting and ending states for motion synthesis.
To this end, we present three synthesized sequences in Fig.~\ref{fig:div}, of which each result leverages three reference subsequences for short-range motion synthesis.
As shown in left part of Fig.~\ref{fig:div}, starting from the same state, we are able to synthesize long-range and visually natural motion, which is facilitated by the short-range and long-range motion generation.

\textcolor{black}
{
\textbf{Reality Evaluation.} 
We evaluate the reality of the synthesized sequences quantitatively. 
We train three classifiers with the ground truth as positive samples. 
Negative samples are generated by three models (i.e., DeepSyn~\cite{DBLP:journals/tog/HoldenSK16}, QuaterNet~\cite{QuaterNet}, and Ours) respectively. 
As shown in Tab.~\ref{tab:cla}, we report the proportion of generated sequences classified as positive ones. 
Row refers to the model used to train the classifier.
Column refers to the model evaluated by the classifiers.
Our model generates more realistic sequences than these two baselines (first two rows).
The proportion of results of the other models classified as positive (third row) is lower.
}

\begin{table}[t!]
\centering
\caption{Classification Accuracy for evaluating the reality of synthesized sequences.}
\begin{tabular}{l | c c c}
\hline
\hline
& DeepSyn~\cite{DBLP:journals/tog/HoldenSK16} & QuaterNet~\cite{QuaterNet} & Ours \\ 
\hline
DeepSyn~\cite{DBLP:journals/tog/HoldenSK16} & --- & 37.6\% & \textbf{66.3\%} \\  
\hline
QuaterNet~\cite{QuaterNet} & 41.9\% & --- & \textbf{75.3\%} \\
\hline
Ours & 28.3\% & 50.4\% & --- \\
\hline
\hline
\end{tabular}
\label{tab:cla}
\end{table}

\textcolor{black}
{
\textbf{Distribution Similarity Evaluation.}
Inspired from FVD score~\cite{DBLP:journals/corr/abs-1812-01717} we measure the distribution similarity between the original and generated sequences with motion features as inputs. The scores (the lower the better) are 281.5 (DeepSyn~\cite{DBLP:journals/tog/HoldenSK16}), 341.1 (QuaterNet~\cite{QuaterNet}), and 179.8 (ours), i.e., the distribution of our results is closer to that of the original data.
Meanwhile, we compare the motion diversity of short-range sequences with those generated by the full model. Motion deviation (the higher the better) is reported, i.e., 0.226 (short-range sequence) and 0.347 (full sequence) respectively, indicating that the long-range generation model facilitates increasing the motion diversity by a large margin.
}

\textbf{Demonstration Guidance for Imitation Learning.}
To further show our model produces realistic motion, generated results are used for demonstration guidance of imitation learning.
As shown in Fig.~\ref{fig:div}, the right part is demonstration synthesized by our model (top) and the bottom one is learned policy with~\cite{peng2018deepmimic}, where the learned motion succeeds in following the synthesized one.

\section{Conclusion}
We present hierarchical style-based networks to generate long-range, diverse and visually plausible motion sequences. Our model trained with large-scale skeleton dataset is also able to generalize to unseen motion much better than previous baselines. We believe our method not only will facilitate graphics applications, but also can be used to generate demonstration for imitation learning. 

\section{Supplementary Material}
\textbf{Model Architecture.} 
Both two encoders consist of 1D convolution layers operating along the temporal axis to capture motion information.
The content encoder consists of two 1D convolution layers.
The first layer is responsible for mapping motion vector to high dimensional space, while the second layer produces content feature.
The style encoder consists of four 1D convolution layers.
The activation function is LeakyReLU for intermediate layers and that of the final layer is Tanh.

\textbf{Design Consideration.}
Note that the motion content changes along with the specific state at each time step, while the style is kept relatively constant throughout the subsequence.
Therefore, to capture instantly changing content feature the kernel width of the two convolutional layers in content encoder are 1 and 3 respectively, i.e., expanding a temporal window with length of 3.
As for the style encoder (4 layers), the kernel width are 11, i.e., the overall receptive field is 40.
With common input frame-rate at 30 fps the style encoder observes about 1.5s of real-world motion, which is sufficient to capture the style feature of one motion sequence. 

\bibliographystyle{splncs04}
\bibliography{egbib}
\end{document}